\documentclass{article}

\PassOptionsToPackage{numbers, compress}{natbib}

\usepackage[final]{workshop_on_video_lm_neurips_2024}

\usepackage[utf8]{inputenc} 
\usepackage[T1]{fontenc}    
\usepackage{hyperref}       
\usepackage{url}            
\usepackage{booktabs}       
\usepackage{amsfonts}       
\usepackage{nicefrac}       
\usepackage{microtype}      
\usepackage{bbm}
\usepackage{graphicx}
\usepackage[dvipsnames]{xcolor}         

\title{HiMemFormer:  Hierarchical Memory-Aware Transformer for Multi-Agent Action Anticipation}

\author{%
Zirui Wang$^1$ \quad Xinran Zhao$^2$ \quad Simon Stepputtis$^2$ \quad Woojun Kim$^2$ \\
\textbf{Tongshuang Wu}$^2$ \quad \textbf{Katia Sycara}$^2$ \quad \textbf{Yaqi Xie}$^2$ \\
$^1$University of Illinois at Urbana-Champaign \quad $^2$Carnegie Mellon University \\
\texttt{ziruiw3@illinois.edu}\\
\texttt{\{xinranz3,sstepput,woojunk,sherryw,sycara,yaqix\}@andrew.cmu.edu}
}

\begin{document}

\maketitle

\begin{abstract}
\looseness = -1
Understanding and predicting human actions has been a long-standing challenge and is a crucial measure of perception in robotics AI. While significant progress has been made in anticipating the future actions of individual agents, prior work has largely overlooked a key aspect of real-world human activity -- interactions. To address this gap in human-like forecasting within multi-agent environments, we present the Hierarchical Memory-Aware Transformer (HiMemFormer), a transformer-based model for online multi-agent action anticipation. HiMemFormer integrates and distributes global memory that captures joint historical information across all agents through a transformer framework, with a hierarchical local memory decoder that interprets agent-specific features based on these global representations using a coarse-to-fine strategy. In contrast to previous approaches, HiMemFormer uniquely hierarchically applies the global context with agent-specific preferences to avoid noisy or redundant information in multi-agent action anticipation. Extensive experiments on various multi-agent scenarios demonstrate the significant performance of HiMemFormer, compared with other state-of-the-art methods. 
\end{abstract}

\section{Introduction}
Action detection \cite{de2016online} or anticipation \cite{kitani2012activity} systems aim at forecasting future states of single or multiple agents from history. The recent advances in these areas facilitate embodied or virtual AI systems with the ability to perceive and interact with other agents and complex environments~\cite{sycara1998multiagent,memvit2022,zhao2022testra}. Such ability plays a pivotal role in numerous applications, such as autonomous driving \cite{yu2020spatio}, collaborative robotics \cite{schydlo2018anticipation}, and home automation \cite{soran2015generating}, where understanding and predicting the actions of various entities in a shared environment can significantly enhance safety, efficiency, and coordination. 

Agent memory plays an important role in conducting action anticipation due to the innate dependencies among actions~\cite{xu2021long, wang2023memory, gong2022future, memvit2022,zhao2022testra}. LSTR~\cite{xu2021long} proposes to capture both long-term and short-term memory, while MAT~\cite{wang2023memory} additionally incorporates future content in seen scenarios. In the multi-agent scenarios~\cite{sycara1998multiagent}, each agent can be arbitrary or affected by the environment,  which suggests one key to the success of a multi-agent system: \textit{how to effectively capture agent behavior at various time and social scales}. A prominent line of research exploits the ways to obtain a unified single global feature representing time, e.g., \cite{kosaraju2019social, alahi2016social}, and social relations, e.g., \cite{huang2019stgat, salzmann2020trajectron++}. AgentFormer~\cite{yuan2021agentformer} and  HiVT~\cite{zhou2022hivt} further explore combining time and social features with a overall global representation. 

Despite the significance, these state-of-the-art systems overlook the individual perspective of the problem: different agents may need time and social features at different scales. From the time perspective,  some agent actions heavily rely on long-term memory, e.g., if they belong to a complex multi-step action sequence, while some actions are only relevant to short memory, e.g., an instant response to a rapid environment change. From the social perspective, similarly, the actions of some agents are much correlated with others during collaboration, while some of mostly stand-alone.
To capture these agent-specific preferences in feature utilization, we propose to \emph{hierarchically} capture the time and social features for each agent-specific decoder to include these global or contextual features with the desired granularity and discard unnecessary information that may introduce noise or latency to \emph{each specific agent}.

To achieve customized and flexible global feature utilization automatically, we propose the Hierarchical Memory-Aware Transformer (HiMemFormer), a novel approach that simultaneously learns feature representations from both contextual and agent-specified dimensions through a dual-hierarchical framework. Specifically, its Agent-to-Context Encoder augments the agents' long-term history through cross-attention with global long-term memory. Then, the encoded long-term memory is further processed through a hierarchical Agent-to-Context Decoder that offers a coarse prediction given augmented long-term memory and contextual short-term memories. Finally, the coarse prediction is gradually refined by each agent-specific network augmented with individual short-term memory to get the anticipated actions. Through the dual-hierarchical network, HiMemFormer manages to model agent's unique short-term memory while learning useful correlations from the contextual memories. This allows us to effectively compress the long-range contextual information without losing important lower level feature information. In summary, our contributions are three-fold:
\begin{itemize}
    \item We propose a transformer-based method to capture and utilize the global features in multi-agent scenarios in a flexible way responding to each agent's preference.
    \item We design a hierarchical memory encoder that follows a specific-to-general paradigm to learn long-term joint-memory features and a hierarchical memory decoder that learns an agent's future action by a coarse-to-fine strategy.
    \item We carry out exhaustive experiments on various multi-agent action anticipation scenarios and outperform existing baseline models. 
\end{itemize}


\section{Hierarchical Memory-Aware Transformer}
\label{himemformer}

\begin{figure}
  \centering
  \includegraphics[clip,trim={0.cm 0.6cm 0.6cm 0.6cm},width=0.95\linewidth]{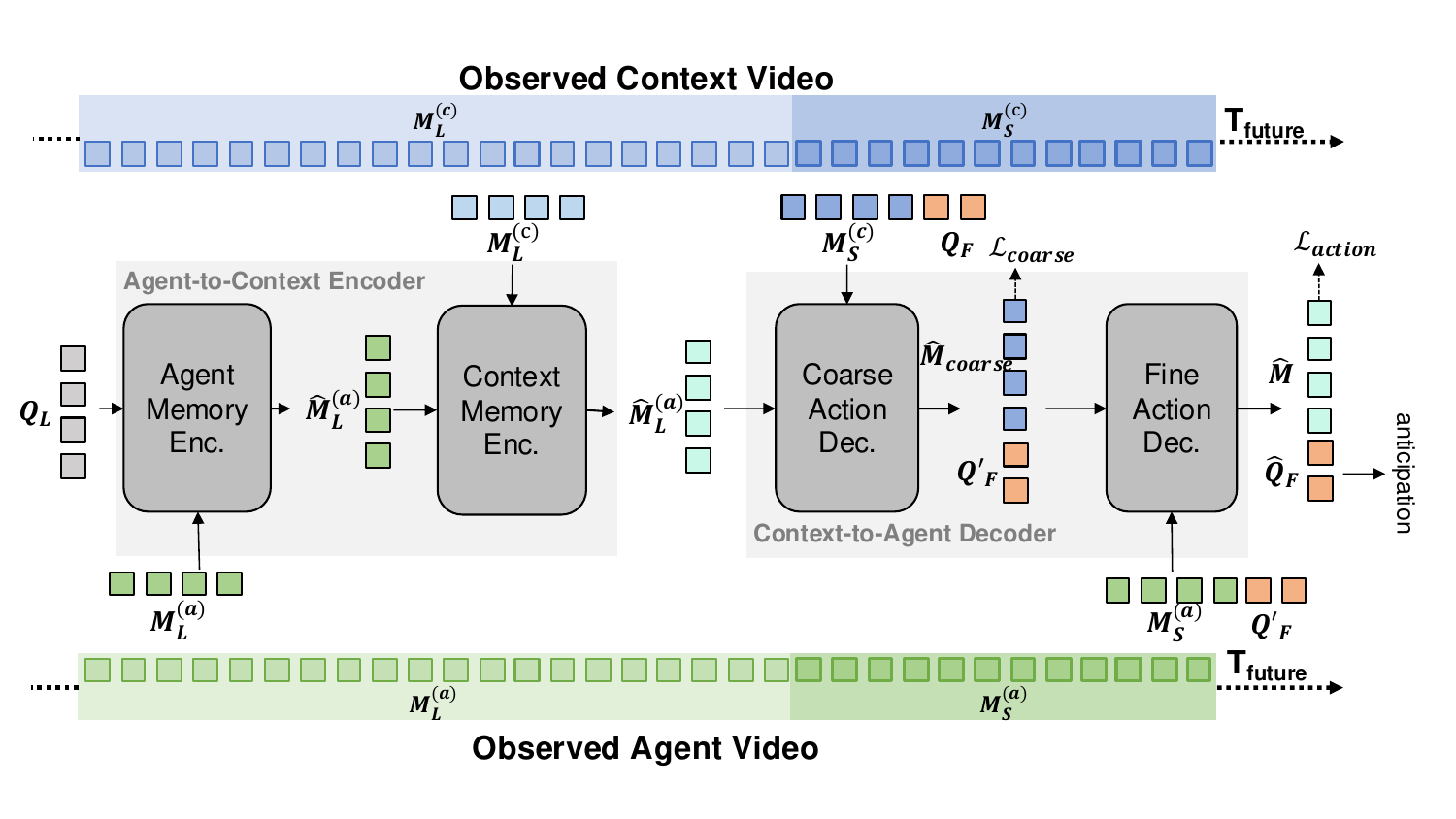}
  \caption{\textbf{HiMemFormer Architecture} In the Agent-to-Context Encoder, the observed agent's long-term memory is encoded to a abstract representation $\hat{\bold{M}}_L^{(a)}$ and cross-attention with context past history $\hat{\bold{M}}_L^{(c)}$. Then, the Context-to-Agent Decoder utilize both agent and global recent memories to learn the future information through a two-stage refinement approach.}
  \label{main_arch}
\end{figure}

We consider the general setting of multi-agent action learning as, given target agent's live streaming First-Person-View (FPV) video, along with the Third-Person-View (TPV) video of the whole scene, our goal is to predict individual agent's actions in a time period $\tau$ using only past and current observations. To tackle this problem, we introduce Hierarchical Memory-Aware Transformer (HiMemFormer), a transformer-based model with encoder-decoder architecture, as shown in Fig. \ref{main_arch}. In particular, given observed agent's long-term history $\bold{M}_L^{(a)}$, we compress it to a latent representation of fixed size through a transformer unit and then cross-attentioned with contextual long-term history $\bold{M}_L^{(c)}$to get the final encoded long-term memory $\bold{M}_L$. Using a coarse-to-fine strategy, we first decode the long-term memory by cross-attentioned with short-term global memory $\bold{M}_S^{(c)}$ to learn all possible actions in the current state, and refine the predicted actions by cross-attention with agent's recent past information $\bold{M}_S^{(a)}$ to get the final anticipated action. See details in Appendix \ref{himemformer_details}


\section{Experiments}
\label{experiment}

\subsection{Datasets and Metrics }
We evaluate our model on a public-available multi-agent dataset LEMMA \cite{jia2020lemma}, which includes 862 compositional atomic-action from 324 activities with 445 egocentric videos (multiple egocentric videos for multi-agent activities). 
We follow prior work \cite{jia2020lemma} on the dataset split, evaluating four scenarios: single-agent single-task $(1 \times 1) $, single-agent multi-tasks $(1 \times 2) $, multi-agent single-task $(2 \times 1) $and multi-agent multi-tasks$(2 \times 2) $.  For online action anticipation, we follow prior works \cite{jia2020lemma,furnari2019would, xu2021long, wang2023memory} and evaluate on per-frame mean average precision (mAP) to measure the performance and evaluate over an anticipation period of $\tau_f = 2s$. See details in Appendix \ref{data_split}.

\subsection{Results and Discussion}

\begin{table}
\small
  \caption{\textbf{Results of online action anticipation on LEMMA} \cite{jia2020lemma} using SlowFast features in up to 2 seconds. In particular, we report accuracy in mAP across 4 scenarios, including single agent scenarios where it perform single or multiple task, and multi-agent scenarios where multiple agents collaborate on single task or carry out separate tasks. 
  }
  \label{main-results}
  \centering
  \begin{tabular}{lllll}
    \toprule
    &                           \multicolumn{4}{c}{Scenarios (\# Agent $\times$ \# Task)}                   \\
    \cmidrule(r){2-5}
                                & $1 \times 1$     & $1 \times 2$ & $2 \times 1$ &  $2 \times 2$ \\
    \midrule
    LSTR\cite{loh2022long}      & 75.8  & 50.9      & 47.0 & 68.0  \\
    MAT \cite{wang2023memory}   & 73.0    &  50.8     &  50.4  & 67.1  \\
    \textbf{HiMemFormer} (ours)  & \textbf{76.3} & \textbf{54.2} & 48.4 & \textbf{70.6} \\
    \textbf{HiMemFormer+} (ours)  & 76.2     &  52.2 & \textbf{50.5} & 69.9 \\
    \bottomrule
  \end{tabular}
\end{table}

We compared HiMemFormer with other baseline models \cite{xu2021long, wang2023memory} on LEMMA \cite{jia2020lemma}. Specifically, we set HiMemFormer with 64 seconds and 5 seconds for long and short-term memories, respectively. Table \ref{main-results} demonstrates that HiMemFormer significantly outperforms LSTR \cite{xu2021long} by at 0.8\%, 4\%, 1.9\% and 0.8\% for all four scenarios respectively in terms of mAP, demonstrating the effectiveness of the hierarchical design of joint agent-specific and contextual memory for inferencing. It is worth noting that HiMemFormer also outperforms MAT \cite{wang2023memory} by a larger margin in $2 \times 2$ scenario, given that MAT additionally learns features from the future. This critical observation indicates the significance of hierarchical global information in multi-agent action anticipation. To ensure a fair comparison, we develop HiMemFormer+ on top of MAT \cite{wang2023memory} to align with its temporal feature on utilizing extra future features, and integrate the hierarchical transformer block for multi-agent action anticipation. Results shows around 2\% improvements over both baselines. More details in Appendix \ref{comparison-with-baselines} and \ref{ablations}.

\section{Conclusion}
We present  Hierarchical Memory-Aware Transformer (HiMemFormer), a transformer-based architecture with hierachical global and local memory attention mechanisms for online action anticipation, to overcome the weakness of the existing methods that can only complete modeling temporal dependency or only modeling agent interaction dependency without considering global historical context. Through experiments on four different scenarios involving multi-agents interactions, we show its capability of modeling both temporal and spatial dependencies, demonstrating the importance of both long-term historical context and short-term agent-specific information.

\section{Acknowledgements and Disclosure of Funding}
This work has been funded in part by the Army Research Laboratory (ARL) award W911NF-23-2-0007, DARPA award FA8750-23-2-1015, and ONR award N00014-23-1-2840. 

\medskip

\bibliography{neurips_2024}

\begin{thebibliography}{45}
\providecommand{\natexlab}[1]{#1}
\providecommand{\url}[1]{\texttt{#1}}
\expandafter\ifx\csname urlstyle\endcsname\relax
  \providecommand{\doi}[1]{doi: #1}\else
  \providecommand{\doi}{doi: \begingroup \urlstyle{rm}\Url}\fi

\bibitem[Abu~Farha and Gall(2019)]{abu2019uncertainty}
Y.~Abu~Farha and J.~Gall.
\newblock Uncertainty-aware anticipation of activities.
\newblock In \emph{Proceedings of the IEEE/CVF International Conference on Computer Vision Workshops}, pages 0--0, 2019.

\bibitem[Abu~Farha et~al.(2018)Abu~Farha, Richard, and Gall]{abu2018will}
Y.~Abu~Farha, A.~Richard, and J.~Gall.
\newblock When will you do what?-anticipating temporal occurrences of activities.
\newblock In \emph{Proceedings of the IEEE conference on computer vision and pattern recognition}, pages 5343--5352, 2018.

\bibitem[Abu~Farha et~al.(2021)Abu~Farha, Ke, Schiele, and Gall]{abu2021long}
Y.~Abu~Farha, Q.~Ke, B.~Schiele, and J.~Gall.
\newblock Long-term anticipation of activities with cycle consistency.
\newblock In \emph{Pattern Recognition: 42nd DAGM German Conference, DAGM GCPR 2020, T{\"u}bingen, Germany, September 28--October 1, 2020, Proceedings 42}, pages 159--173. Springer, 2021.

\bibitem[Alahi et~al.(2016)Alahi, Goel, Ramanathan, Robicquet, Fei-Fei, and Savarese]{alahi2016social}
A.~Alahi, K.~Goel, V.~Ramanathan, A.~Robicquet, L.~Fei-Fei, and S.~Savarese.
\newblock Social lstm: Human trajectory prediction in crowded spaces.
\newblock In \emph{Proceedings of the IEEE conference on computer vision and pattern recognition}, pages 961--971, 2016.

\bibitem[Bhagat et~al.(2023)Bhagat, Stepputtis, Campbell, and Sycara]{bhagat2023knowledgeguidedshortcontextactionanticipation}
S.~Bhagat, S.~Stepputtis, J.~Campbell, and K.~Sycara.
\newblock Knowledge-guided short-context action anticipation in human-centric videos, 2023.
\newblock URL \url{https://arxiv.org/abs/2309.05943}.

\bibitem[Chen et~al.(2024)Chen, Lv, Wu, Lin, Song, Gao, Liu, Gao, Mao, and Shou]{chen2024videollm}
J.~Chen, Z.~Lv, S.~Wu, K.~Q. Lin, C.~Song, D.~Gao, J.-W. Liu, Z.~Gao, D.~Mao, and M.~Z. Shou.
\newblock Videollm-online: Online video large language model for streaming video.
\newblock In \emph{Proceedings of the IEEE/CVF Conference on Computer Vision and Pattern Recognition}, pages 18407--18418, 2024.

\bibitem[Damen et~al.(2022)Damen, Doughty, Farinella, Furnari, Ma, Kazakos, Moltisanti, Munro, Perrett, Price, and Wray]{Damen2022RESCALING}
D.~Damen, H.~Doughty, G.~M. Farinella, A.~Furnari, J.~Ma, E.~Kazakos, D.~Moltisanti, J.~Munro, T.~Perrett, W.~Price, and M.~Wray.
\newblock Rescaling egocentric vision: Collection, pipeline and challenges for epic-kitchens-100.
\newblock \emph{International Journal of Computer Vision (IJCV)}, 130:\penalty0 33–55, 2022.
\newblock URL \url{https://doi.org/10.1007/s11263-021-01531-2}.

\bibitem[De~Geest et~al.(2016)De~Geest, Gavves, Ghodrati, Li, Snoek, and Tuytelaars]{de2016online}
R.~De~Geest, E.~Gavves, A.~Ghodrati, Z.~Li, C.~Snoek, and T.~Tuytelaars.
\newblock Online action detection.
\newblock In \emph{Computer Vision--ECCV 2016: 14th European Conference, Amsterdam, The Netherlands, October 11-14, 2016, Proceedings, Part V 14}, pages 269--284. Springer, 2016.

\bibitem[Fan et~al.(2020)Fan, Li, Xiong, Lo, and Feichtenhofer]{fan2020pyslowfast}
H.~Fan, Y.~Li, B.~Xiong, W.-Y. Lo, and C.~Feichtenhofer.
\newblock Pyslowfast.
\newblock \url{https://github.com/facebookresearch/slowfast}, 2020.

\bibitem[Furnari and Farinella(2019)]{furnari2019would}
A.~Furnari and G.~M. Farinella.
\newblock What would you expect? anticipating egocentric actions with rolling-unrolling lstms and modality attention.
\newblock In \emph{Proceedings of the IEEE/CVF International conference on computer vision}, pages 6252--6261, 2019.

\bibitem[Furnari et~al.(2018)Furnari, Battiato, and Maria~Farinella]{furnari2018leveraging}
A.~Furnari, S.~Battiato, and G.~Maria~Farinella.
\newblock Leveraging uncertainty to rethink loss functions and evaluation measures for egocentric action anticipation.
\newblock In \emph{Proceedings of the European conference on computer vision (ECCV) workshops}, pages 0--0, 2018.

\bibitem[Gammulle et~al.(2019)Gammulle, Denman, Sridharan, and Fookes]{gammulle2019predicting}
H.~Gammulle, S.~Denman, S.~Sridharan, and C.~Fookes.
\newblock Predicting the future: A jointly learnt model for action anticipation.
\newblock In \emph{Proceedings of the IEEE/CVF International Conference on Computer Vision}, pages 5562--5571, 2019.

\bibitem[Girase et~al.(2023)Girase, Agarwal, Choi, and Mangalam]{girase2023latency}
H.~Girase, N.~Agarwal, C.~Choi, and K.~Mangalam.
\newblock Latency matters: Real-time action forecasting transformer.
\newblock In \emph{Proceedings of the IEEE/CVF Conference on Computer Vision and Pattern Recognition}, pages 18759--18769, 2023.

\bibitem[Gong et~al.(2022)Gong, Lee, Kim, Ha, and Cho]{gong2022future}
D.~Gong, J.~Lee, M.~Kim, S.~J. Ha, and M.~Cho.
\newblock Future transformer for long-term action anticipation.
\newblock In \emph{Proceedings of the IEEE/CVF Conference on Computer Vision and Pattern Recognition}, pages 3052--3061, 2022.

\bibitem[Grauman et~al.(2022)Grauman, Westbury, Byrne, Chavis, Furnari, Girdhar, Hamburger, Jiang, Liu, Liu, Martin, Nagarajan, Radosavovic, Ramakrishnan, Ryan, Sharma, Wray, Xu, Xu, Zhao, Bansal, Batra, Cartillier, Crane, Do, Doulaty, Erapalli, Feichtenhofer, Fragomeni, Fu, Fuegen, Gebreselasie, Gonzalez, Hillis, Huang, Huang, Jia, Khoo, Kolar, Kottur, Kumar, Landini, Li, Li, Li, Mangalam, Modhugu, Munro, Murrell, Nishiyasu, Price, Puentes, Ramazanova, Sari, Somasundaram, Southerland, Sugano, Tao, Vo, Wang, Wu, Yagi, Zhu, Arbelaez, Crandall, Damen, Farinella, Ghanem, Ithapu, Jawahar, Joo, Kitani, Li, Newcombe, Oliva, Park, Rehg, Sato, Shi, Shou, Torralba, Torresani, Yan, and Malik]{Ego4D2022CVPR}
K.~Grauman, A.~Westbury, E.~Byrne, Z.~Chavis, A.~Furnari, R.~Girdhar, J.~Hamburger, H.~Jiang, M.~Liu, X.~Liu, M.~Martin, T.~Nagarajan, I.~Radosavovic, S.~K. Ramakrishnan, F.~Ryan, J.~Sharma, M.~Wray, M.~Xu, E.~Z. Xu, C.~Zhao, S.~Bansal, D.~Batra, V.~Cartillier, S.~Crane, T.~Do, M.~Doulaty, A.~Erapalli, C.~Feichtenhofer, A.~Fragomeni, Q.~Fu, C.~Fuegen, A.~Gebreselasie, C.~Gonzalez, J.~Hillis, X.~Huang, Y.~Huang, W.~Jia, W.~Khoo, J.~Kolar, S.~Kottur, A.~Kumar, F.~Landini, C.~Li, Y.~Li, Z.~Li, K.~Mangalam, R.~Modhugu, J.~Munro, T.~Murrell, T.~Nishiyasu, W.~Price, P.~R. Puentes, M.~Ramazanova, L.~Sari, K.~Somasundaram, A.~Southerland, Y.~Sugano, R.~Tao, M.~Vo, Y.~Wang, X.~Wu, T.~Yagi, Y.~Zhu, P.~Arbelaez, D.~Crandall, D.~Damen, G.~M. Farinella, B.~Ghanem, V.~K. Ithapu, C.~V. Jawahar, H.~Joo, K.~Kitani, H.~Li, R.~Newcombe, A.~Oliva, H.~S. Park, J.~M. Rehg, Y.~Sato, J.~Shi, M.~Z. Shou, A.~Torralba, L.~Torresani, M.~Yan, and J.~Malik.
\newblock Ego4d: Around the {W}orld in 3,000 {H}ours of {E}gocentric {V}ideo.
\newblock In \emph{IEEE/CVF Computer Vision and Pattern Recognition (CVPR)}, 2022.

\bibitem[Hogan et~al.(2021)Hogan, Blomqvist, Cochez, d’Amato, Melo, Gutierrez, Kirrane, Gayo, Navigli, Neumaier, et~al.]{hogan2021knowledge}
A.~Hogan, E.~Blomqvist, M.~Cochez, C.~d’Amato, G.~D. Melo, C.~Gutierrez, S.~Kirrane, J.~E.~L. Gayo, R.~Navigli, S.~Neumaier, et~al.
\newblock Knowledge graphs.
\newblock \emph{ACM Computing Surveys (Csur)}, 54\penalty0 (4):\penalty0 1--37, 2021.

\bibitem[Huang et~al.(2019)Huang, Bi, Li, Mao, and Wang]{huang2019stgat}
Y.~Huang, H.~Bi, Z.~Li, T.~Mao, and Z.~Wang.
\newblock Stgat: Modeling spatial-temporal interactions for human trajectory prediction.
\newblock In \emph{Proceedings of the IEEE/CVF international conference on computer vision}, pages 6272--6281, 2019.

\bibitem[Huang et~al.(2021)Huang, Yang, and Xu]{10.1145/3474085.3475327}
Y.~Huang, X.~Yang, and C.~Xu.
\newblock Multimodal global relation knowledge distillation for egocentric action anticipation.
\newblock In \emph{Proceedings of the 29th ACM International Conference on Multimedia}, MM '21, page 245–254, New York, NY, USA, 2021. Association for Computing Machinery.
\newblock ISBN 9781450386517.
\newblock \doi{10.1145/3474085.3475327}.
\newblock URL \url{https://doi.org/10.1145/3474085.3475327}.

\bibitem[Jain et~al.(2016)Jain, Singh, Koppula, Soh, and Saxena]{jain2016recurrent}
A.~Jain, A.~Singh, H.~S. Koppula, S.~Soh, and A.~Saxena.
\newblock Recurrent neural networks for driver activity anticipation via sensory-fusion architecture.
\newblock In \emph{2016 IEEE international conference on robotics and automation (ICRA)}, pages 3118--3125. IEEE, 2016.

\bibitem[Jia et~al.(2020)Jia, Chen, Huang, Zhu, and Zhu]{jia2020lemma}
B.~Jia, Y.~Chen, S.~Huang, Y.~Zhu, and S.-C. Zhu.
\newblock Lemma: A multiview dataset for learning multi-agent multi-view activities.
\newblock In \emph{Proceedings of the European Conference on Computer Vision (ECCV)}, 2020.

\bibitem[Kitani et~al.(2012)Kitani, Ziebart, Bagnell, and Hebert]{kitani2012activity}
K.~M. Kitani, B.~D. Ziebart, J.~A. Bagnell, and M.~Hebert.
\newblock Activity forecasting.
\newblock In \emph{Computer Vision--ECCV 2012: 12th European Conference on Computer Vision, Florence, Italy, October 7-13, 2012, Proceedings, Part IV 12}, pages 201--214. Springer, 2012.

\bibitem[Kong et~al.(2018)Kong, Gao, Sun, and Fu]{kong2018action}
Y.~Kong, S.~Gao, B.~Sun, and Y.~Fu.
\newblock Action prediction from videos via memorizing hard-to-predict samples.
\newblock In \emph{Proceedings of the AAAI conference on artificial intelligence}, volume~32, 2018.

\bibitem[Kosaraju et~al.(2019)Kosaraju, Sadeghian, Mart{\'\i}n-Mart{\'\i}n, Reid, Rezatofighi, and Savarese]{kosaraju2019social}
V.~Kosaraju, A.~Sadeghian, R.~Mart{\'\i}n-Mart{\'\i}n, I.~Reid, H.~Rezatofighi, and S.~Savarese.
\newblock Social-bigat: Multimodal trajectory forecasting using bicycle-gan and graph attention networks.
\newblock \emph{Advances in neural information processing systems}, 32, 2019.

\bibitem[Li et~al.(2021)Li, Liu, and Rehg]{li2021eye}
Y.~Li, M.~Liu, and J.~M. Rehg.
\newblock In the eye of the beholder: Gaze and actions in first person video.
\newblock \emph{IEEE transactions on pattern analysis and machine intelligence}, 45\penalty0 (6):\penalty0 6731--6747, 2021.

\bibitem[Li et~al.(2022)Li, Wu, Fan, Mangalam, Xiong, Malik, and Feichtenhofer]{li2022mvitv2}
Y.~Li, C.-Y. Wu, H.~Fan, K.~Mangalam, B.~Xiong, J.~Malik, and C.~Feichtenhofer.
\newblock Mvitv2: Improved multiscale vision transformers for classification and detection.
\newblock In \emph{Proceedings of the IEEE/CVF conference on computer vision and pattern recognition}, pages 4804--4814, 2022.

\bibitem[Loh et~al.(2022)Loh, Roy, and Fernando]{loh2022long}
S.~B. Loh, D.~Roy, and B.~Fernando.
\newblock Long-term action forecasting using multi-headed attention-based variational recurrent neural networks.
\newblock In \emph{Proceedings of the IEEE/CVF Conference on Computer Vision and Pattern Recognition}, pages 2419--2427, 2022.

\bibitem[Rodriguez et~al.(2018)Rodriguez, Fernando, and Li]{rodriguez2018action}
C.~Rodriguez, B.~Fernando, and H.~Li.
\newblock Action anticipation by predicting future dynamic images.
\newblock In \emph{Proceedings of the European Conference on Computer Vision (ECCV) Workshops}, pages 0--0, 2018.

\bibitem[Salzmann et~al.(2020)Salzmann, Ivanovic, Chakravarty, and Pavone]{salzmann2020trajectron++}
T.~Salzmann, B.~Ivanovic, P.~Chakravarty, and M.~Pavone.
\newblock Trajectron++: Dynamically-feasible trajectory forecasting with heterogeneous data.
\newblock In \emph{Computer Vision--ECCV 2020: 16th European Conference, Glasgow, UK, August 23--28, 2020, Proceedings, Part XVIII 16}, pages 683--700. Springer, 2020.

\bibitem[Schydlo et~al.(2018)Schydlo, Rakovic, Jamone, and Santos-Victor]{schydlo2018anticipation}
P.~Schydlo, M.~Rakovic, L.~Jamone, and J.~Santos-Victor.
\newblock Anticipation in human-robot cooperation: A recurrent neural network approach for multiple action sequences prediction.
\newblock In \emph{2018 IEEE International Conference on Robotics and Automation (ICRA)}, pages 5909--5914. IEEE, 2018.

\bibitem[Sener et~al.(2020)Sener, Singhania, and Yao]{sener2020temporal}
F.~Sener, D.~Singhania, and A.~Yao.
\newblock Temporal aggregate representations for long-range video understanding.
\newblock In \emph{Computer Vision--ECCV 2020: 16th European Conference, Glasgow, UK, August 23--28, 2020, Proceedings, Part XVI 16}, pages 154--171. Springer, 2020.

\bibitem[Soran et~al.(2015)Soran, Farhadi, and Shapiro]{soran2015generating}
B.~Soran, A.~Farhadi, and L.~Shapiro.
\newblock Generating notifications for missing actions: Don't forget to turn the lights off!
\newblock In \emph{Proceedings of the IEEE International Conference on Computer Vision}, pages 4669--4677, 2015.

\bibitem[Sycara(1998)]{sycara1998multiagent}
K.~P. Sycara.
\newblock Multiagent systems.
\newblock \emph{AI magazine}, 19\penalty0 (2):\penalty0 79--79, 1998.

\bibitem[Tong et~al.(2022)Tong, Song, Wang, and Wang]{tong2022videomae}
Z.~Tong, Y.~Song, J.~Wang, and L.~Wang.
\newblock Videomae: Masked autoencoders are data-efficient learners for self-supervised video pre-training.
\newblock \emph{Advances in neural information processing systems}, 35:\penalty0 10078--10093, 2022.

\bibitem[Vaswani et~al.(2017)Vaswani, Shazeer, Parmar, Uszkoreit, Jones, Gomez, Kaiser, and Polosukhin]{vaswani2017attention}
A.~Vaswani, N.~Shazeer, N.~Parmar, J.~Uszkoreit, L.~Jones, A.~N. Gomez, {\L}.~Kaiser, and I.~Polosukhin.
\newblock Attention is all you need.
\newblock \emph{Advances in neural information processing systems}, 30, 2017.

\bibitem[Wang et~al.(2023)Wang, Chen, Huang, Wang, and Lu]{wang2023memory}
J.~Wang, G.~Chen, Y.~Huang, L.~Wang, and T.~Lu.
\newblock Memory-and-anticipation transformer for online action understanding.
\newblock In \emph{Proceedings of the IEEE/CVF International Conference on Computer Vision}, pages 13824--13835, 2023.

\bibitem[Wu et~al.(2022)Wu, Li, Mangalam, Fan, Xiong, Malik, and Feichtenhofer]{memvit2022}
C.-Y. Wu, Y.~Li, K.~Mangalam, H.~Fan, B.~Xiong, J.~Malik, and C.~Feichtenhofer.
\newblock {MeMViT: Memory-Augmented Multiscale Vision Transformer for Efficient Long-Term Video Recognition}.
\newblock In \emph{CVPR}, 2022.

\bibitem[Xu et~al.(2017)Xu, Zhu, Choy, and Fei-Fei]{xu2017scene}
D.~Xu, Y.~Zhu, C.~B. Choy, and L.~Fei-Fei.
\newblock Scene graph generation by iterative message passing.
\newblock In \emph{Proceedings of the IEEE conference on computer vision and pattern recognition}, pages 5410--5419, 2017.

\bibitem[Xu et~al.(2021)Xu, Xiong, Chen, Li, Xia, Tu, and Soatto]{xu2021long}
M.~Xu, Y.~Xiong, H.~Chen, X.~Li, W.~Xia, Z.~Tu, and S.~Soatto.
\newblock Long short-term transformer for online action detection.
\newblock In \emph{Conference on Neural Information Processing Systems (NeurIPS)}, 2021.

\bibitem[Yang et~al.(2022)Yang, Han, and Zhang]{yang2022colar}
L.~Yang, J.~Han, and D.~Zhang.
\newblock Colar: Effective and efficient online action detection by consulting exemplars.
\newblock In \emph{Proceedings of the IEEE/CVF conference on computer vision and pattern recognition}, pages 3160--3169, 2022.

\bibitem[Yu et~al.(2020)Yu, Ma, Ren, Zhao, and Yi]{yu2020spatio}
C.~Yu, X.~Ma, J.~Ren, H.~Zhao, and S.~Yi.
\newblock Spatio-temporal graph transformer networks for pedestrian trajectory prediction.
\newblock In \emph{Computer Vision--ECCV 2020: 16th European Conference, Glasgow, UK, August 23--28, 2020, Proceedings, Part XII 16}, pages 507--523. Springer, 2020.

\bibitem[Yuan et~al.(2021)Yuan, Weng, Ou, and Kitani]{yuan2021agentformer}
Y.~Yuan, X.~Weng, Y.~Ou, and K.~M. Kitani.
\newblock Agentformer: Agent-aware transformers for socio-temporal multi-agent forecasting.
\newblock In \emph{Proceedings of the IEEE/CVF International Conference on Computer Vision}, pages 9813--9823, 2021.

\bibitem[Zhao et~al.(2023)Zhao, Zhang, Wang, Fu, Agarwal, Lee, and Sun]{zhao2023antgpt}
Q.~Zhao, C.~Zhang, S.~Wang, C.~Fu, N.~Agarwal, K.~Lee, and C.~Sun.
\newblock Antgpt: Can large language models help long-term action anticipation from videos?
\newblock \emph{arXiv preprint arXiv:2307.16368}, 2023.

\bibitem[Zhao and Kr{\"a}henb{\"u}hl(2022)]{zhao2022testra}
Y.~Zhao and P.~Kr{\"a}henb{\"u}hl.
\newblock Real-time online video detection with temporal smoothing transformers.
\newblock In \emph{European Conference on Computer Vision (ECCV)}, 2022.

\bibitem[Zhou et~al.(2022)Zhou, Ye, Wang, Wu, and Lu]{zhou2022hivt}
Z.~Zhou, L.~Ye, J.~Wang, K.~Wu, and K.~Lu.
\newblock Hivt: Hierarchical vector transformer for multi-agent motion prediction.
\newblock In \emph{Proceedings of the IEEE/CVF Conference on Computer Vision and Pattern Recognition}, pages 8823--8833, 2022.

\bibitem[Zou et~al.(2024)Zou, Zhou, Li, Han, and Zhang]{zou2024promptintern}
J.~Zou, M.~Zhou, T.~Li, S.~Han, and D.~Zhang.
\newblock Promptintern: Saving inference costs by internalizing recurrent prompt during large language model fine-tuning.
\newblock \emph{arXiv preprint arXiv:2407.02211}, 2024.

\end{thebibliography}
\bibliographystyle{abbrvnat}


\appendix

\section{Related Work}
\paragraph{Action Anticipation}
Online action anticipation \cite{kitani2012activity} aims to predict the future actions of the agent given the past and current action information. Given its' increasing popularity and its broad practical applications, many large-scale datasets and benchmarks  \cite{Damen2022RESCALING, Ego4D2022CVPR, li2021eye} has been proposed to facilitate researchers in this area. In the field of action anticipation, feature learning \cite{rodriguez2018action, gammulle2019predicting} and temporal modeling \cite{abu2018will, gong2022future} are the two main streams of approaches. Recurrent neural network (RNN) is widely adopted by many prior works \cite{abu2019uncertainty, abu2021long, gammulle2019predicting, kong2018action, loh2022long, furnari2019would} due to its powerful long-term temporal dependency. For example, RULSTM \cite{furnari2019would} proposed to anticipate actions via a rolling LSTM to encode historical information and unrolling LSTM make predictions on future actions. To better model time and social dimension in a multi-agent setting, a popular line of research \cite{kosaraju2019social, alahi2016social} uses temporal models to summarize features over time for each agent separately and then feed temporal features into social models to obtain global-aware agent features. There are also works \cite{huang2019stgat, salzmann2020trajectron++} that uses social models to generate social features of for individual agents and apply temporal models to summarize social features for each agent. For example, Trajection++ \cite{salzmann2020trajectron++} design a graph-structured recurrent model that forecasts the trajectories of a general number of diverse agents while incorporating agent dynamics and heterogeneous data. However, previous methods failed to consider both temporal dependencies and social dependencies at once, which can be sub-optimal. More recent work \cite{yuan2021agentformer, zhou2022hivt} manage to overcome this short coming by considering both time and social dimensions simultaneously, facilitating interaction across temporal domain and spatial domain. 
\paragraph{Transformer for Video Understanding} Recently, transformer-based methods \cite{gong2022future, wang2023memory, xu2021long, li2022mvitv2,tong2022videomae, yang2022colar,girase2023latency} stood out in literature because of its strong capability for long-range temporal dependencies. For example, Gong \textit{et al.} \cite{gong2022future} proposed Future Transformer (FUTR), an end-to-end attention neural network that anticipate actions in parallel decoding, leveraging global interactions between past and future actions for long-term action anticipation. LSTR \cite{xu2021long} further decomposes the memory encoder into long and short-term stages for online action detection and anticipation, allowing model to learn more representative features from the history. MAT \cite{wang2023memory} proposes a new memory-anticipation-based paradigm that models the entire temporal structure, including past, present and future. Also to utilize different sources such as optical flow and audio data, multi-modal fusion approaches \cite{furnari2019would, furnari2018leveraging, jain2016recurrent, sener2020temporal} has been proposed to improve the accuracy of future action prediction. In addition, large language model (LLM) \cite{zhao2023antgpt} is deployed to tackle action anticipation task due to its strong high-level reasoning capability.

\section{HiMemFormer Details}
\label{himemformer_details}
\subsection{Agent-to-Context Encoder}
Agent-specific long-term memory provide useful information about the historical actions of the agent, but when placed in a complex environment with multi-agent interactions, it is crucial to pay attention to contextual information that are shared across all agents. To this end, Agent-to-Context Memory Encoder follows the specific-to-general approach, managing to augment agent's long-term memory by paying extra attention to global features via cross-attention. 

\paragraph{Agent Memory Encoding.} For each agent in a scene, we input target agent's long-term memory features $\bold{M}_L^{(a)}$ to the Transformer Block and compress target agent's long-term feature into a latent representation of fixed length. Following prior work \cite{xu2021long, zhao2022testra}, we utilize a two-stage memory compression transformer module, denoted $\mathcal{F}_L^{(a)}$, consists of multiple transformer decoder unit \cite{vaswani2017attention} to encoder the agent-specific long-term history $\widehat{\bold{M}}_L^{(a)}$:
\begin{equation}
    \widehat{\bold{M}}_L^{(a)} = \mathcal{F}_L^{(a)}(\bold{M}_L^{(a)}, \bold{M}_L^{(a)})
\end{equation}

\paragraph{Context Memory Enhancement.} To effectively encode contextual information into agent's long-term memory, we propose a specific-to-general approach. In practice, we send contextual long-term history $\bold{M}_L^{(c)}$ (with positional embedding) as queries and $\widehat{\bold{M}}_L^{(a)}$ to our context encoder, $\mathcal{F}_L^{(c)}$, constructed with a transformer decoder architecture \cite{vaswani2017attention}. Using contextual long-term history to guide the encoded agent's long-term history, we acquire the final encoded long-term memory $\widehat{\bold{M}}_L$:
\begin{equation}
    \widehat{\bold{M}}_L = \mathcal{F}_L^{(c)}(\bold{M}_L^{(c)}, \widehat{\bold{M}}_L^{(a)})
\end{equation}
where inputs to the decoder $\mathcal{F}$ are queries, and key/value pairs. 
\subsection{Context-to-Agent Decoder}
To implement our key idea to predict agent future actions based on both contextual information and agent-specific information, we meticulously design our Context-to-Agent Decoder using a coarse-to-fine apporach. In particular, leveraging informative short-term features, as demonstrated in LSTR \cite{xu2021long}, we first make a coarse prediction that contains possible future actions of all agents in the scene using contextual short-term features, and narrow down to target agent's future action using agent-specific short-term features as queries to the transformers units. We also supervise on both coarse and precise action predictions.

\paragraph{Coarse Action Anticipation} To enable the model to learn future actions, we need to generate a latent embedding that allows the model to learn future actions from the past and the present. In practise, we initialize $N_F$ learnable query tokens $\bold{Q}_F \in \mathbb{R}^{N_F\times D}$ where $D$ is the feature dimension and concatenate with contextual short-term memories $\bold{M}_S^{(c)}$ to form $\bold{M}_{coarse} = \{ \bold{M}_S^{(c)},  \bold{Q}_F^{(c)}\}$. We then take $\bold{M}_{coarse}^{(c)}$ as queries and cross-attentioned with augmented long-term memory $\bold{M}_L$ through a transformer decoder architecture \cite{vaswani2017attention} $\mathcal{F}_{coarse}$ to make coarse action predictions given only contextual information:
\begin{equation}
     \widehat{\bold{M}}_{coarse} = \mathcal{F}_{coarse}(\bold{M}_{coarse}, \widehat{\bold{M}}_L)
\end{equation}

\paragraph{Precise Action Refinement} Until now, the future action query token contain general information about agent's future action. To generate more accurate predicted actions, we leverage the agent-specific short-term history, $\bold{M}_S^{(a)}$, that contains feature representations of agent's unique feature and concatenate them with learnable query tokens from the coarse prediction $\bold{Q}'_F$ to form $\bold{M}_{fine} = \{ \bold{M}_S^{(a)},  \bold{Q}'_F\}$. Following the similar recipe we add another transformer block, denoted $\mathcal{F}_{fine}$ to generate the final action prediction $\widehat{\bold{M}}$:
\begin{equation}
     \widehat{\bold{M}} = \mathcal{F}_{fine}(\bold{M}_{fine},  \widehat{\bold{M}}_{coarse})
\end{equation}

\subsection{Training Objectives}
The loss function for HiMemFormer comprises two essential components: the coarse action loss $\mathcal{L}_{coarse}$ and the refined action loss $\mathcal{L}_{fine}$. The overall respresentation is:
\begin{equation}
    \mathcal{L} = \lambda_a \cdot \mathcal{L}_{coarse}+ \lambda_b \cdot \mathcal{L}_{fine}
\end{equation}
We then use the empirical cross entropy loss between each agent's predicted action anticipation probability distribution $\hat{P}_t \in \mathbb{R}^{T \times (K + 1)}$ and the ground truth anticipation label $y_t \in \{0, 1,...,K\}$. For the coarse action anticipation, we utilize the ground truth anticipation label of all agents in the scene. For the refined action anticipation, we ony use the ground truth anticipation label of the target agent. 

\section{Experiments}

\subsection{Data Splits}
\label{data_split}

We randomly split all the video samples into training and test sets with ratio of 3 to 1, resulting in 243 recorded activities for training and 81 for validation. Due to the multi-agent setup, the model will be trained on 333 out of 445 egocentric videos, as each activity may have multiple egocentric videos. For action anticipation task, we follow prior work and split training and validation sets with ratios 3: 1, 1:3, 1: 3 and 1:3 for the four scenarios $1 \times 1, 1 \times 2, 2 \times 1, 2 \times 2$, respectively, resulting in (96,19,16,13) activities for training and (31, 57, 50, 42) activities for evaluation in four scenarios. 

\subsection{Experiment Settings}
We implemented our proposed model in PyTorch and performed all experiments on a system with a  single NVIDIA A40 graphics cards. For all transformer blocks inside both encoder and decoder module, we set the number of heads to 4 and hidden units as 1024 dimensions. The model is optimized by AdamW optimizer with a weight decay of $1 \times 10^{-4}$ We use warm-up learning rate linearly increase from zero to $7 \times 10^{-5}$ in the first 10 epoch. In addition, the model is optimized with batch size of 16 and training is terminated after 25 epochs. Following prior work's experiment settings, we use a pretrained feature extractor \cite{fan2020pyslowfast} extract action features from the video. 

\subsection{Comparison with Baselines}
\label{comparison-with-baselines}
We compare HiMemFormer with prior methods on LEMMA for action anticipation in both single- and multi-agent environment. Specifically, both baselines, LSTR \cite{xu2021long} and MAT \cite{wang2023memory}, only take in agent's first-person-view live streaming video as the input without considering the contextual information from the third-person-view videos. The above set up ensure that the performance gain shown in table \ref{main-results} come from the proposed multi-view integration of agent and context memory.

\subsection{Ablation Studies}
\label{ablations}
\begin{table}
\small
\caption{\textbf{Exploring different short-term memory size} of HiMemFormer. In particular, we fixed the long-term memory size to 64 seconds. $M_S$ represents length of short-term memory (in secs).
  }
  \label{table-short-memory-size}
  \centering
  \begin{tabular}{lllll}
    \toprule
    &                           \multicolumn{4}{c}{Scenarios (\# Agent $\times$ \# Task)}                   \\
    \cmidrule(r){2-5}
     & $1 \times 1$     & $1 \times 2$ & $2 \times 1$ &  $2 \times 2$ \\
    \midrule
    $M_S = 2$       & 73.5  & 52.1      & 47.8 & \textbf{73.4}  \\
    $M_S = 5$   & \textbf{76.3}    &  \textbf{54.2}  &  \textbf{48.4}  & 70.6  \\
    $M_S = 10$  & 76.1  & 47.5   & 45.4& 68.3 \\
    \bottomrule
  \end{tabular}
\end{table}

\begin{table}
\small
\caption{\textbf{Exploring different long-term memory size} of HiMemFormer. In particular, we fixed the short-term memory size to 5 seconds. $M_L$ represents length of long-term memory (in secs).
  }
  \label{table-long-memory-size}
  \centering
  \begin{tabular}{lllll}
    \toprule
    &                           \multicolumn{4}{c}{Scenarios (\# Agent $\times$ \# Task)}                   \\
    \cmidrule(r){2-5}
     & $1 \times 1$     & $1 \times 2$ & $2 \times 1$ &  $2 \times 2$ \\
    \midrule
    $M_L = 32$  & 76.3	& 53.7	& 47.7	& 70  \\
    $M_L = 64$  & 76.3 & \textbf{54.2} & \textbf{48.4} & \textbf{70.6} \\
    $M_L = 128$  & \textbf{77.4}	& 51.2	& 45.4 & 68.2 \\
    $M_L = 256$  & 74.5	& 48.9 & 46.4 & 68.7 \\
    \bottomrule
  \end{tabular}
\end{table}

\begin{table}
\small
  \caption{\textbf{Effect of different down-sampling rate for long-term memory} \cite{jia2020lemma} on LEMMA \cite{jia2020lemma}. In particularly, we implement HiMemForMer with long-term memory size of 64 seconds and short-term memory size of 5 seconds.}
  \label{table-sampling-rate}
  \centering
  \begin{tabular}{lllll}
    \toprule
    &                           \multicolumn{4}{c}{Scenarios (\# Agent $\times$ \# Task)}                   \\
    \cmidrule(r){2-5}
    Sampling Rate (SR) & $1 \times 1$     & $1 \times 2$ & $2 \times 1$ &  $2 \times 2$ \\
    \midrule
    $SR = 1 $     & 77.4& 52.7 & 49.3 & 69.7 \\
    $SR = 4$ & 76.3 & 54.2 & 48.4 & 70.6 \\
    $SR = 8 $ & 76.3 & 53.3 & 49.2 & 69.6 \\
    $SR = 16$ & 73.0 & 52.3 & 47.0 & 67.6 \\
    \bottomrule
  \end{tabular}
\end{table}

\paragraph{Effect of Memory Size} We first analyze the effect of the length of both short term and long term memory. Following LSTR \cite{xu2021long}, we fix the short-term memory to 5 seconds and test $M_L \in \{32, 64, 128, 256 \}$, while maintaining same memory size for multi-view videos. Similarly, we fix the long-term memory to 64 seconds and test $M_S \in \{ 2, 5,10\}$. Results shown in table \ref{table-short-memory-size} and \ref{table-long-memory-size} reach the same conclusion in \cite{xu2021long}, where increasing memory size does not always guarantee better performance. 
\paragraph{Effect of Down-Sampling Rate} We also test the effect of compression ratio of long-term memory, and we implement HiMemFormer with 5 seconds of short-term memories (both agent and global perspective) and 64 seconds of long-term memories. Results are shown in table \ref{table-sampling-rate}.

\subsection{Discussion on Future Directions}
HiMemFormer serves as an attempt to tackle action-anticipation in complex multi-agent environment, but there's more to be explored. Future work could focus on expanding HiMemFormer’s capabilities to better interpret complex multi-agent interactions. Potential directions include leveraging large language models (LLMs) to enhance model's interpretability and flexibility in the dynamic environment \cite{zhao2023antgpt, chen2024videollm, zou2024promptintern}. Knowledge graphs \cite{hogan2021knowledge} or scene graphs \cite{xu2017scene} can also serves as a powerful feature representations \cite{bhagat2023knowledgeguidedshortcontextactionanticipation, 10.1145/3474085.3475327}to further boost the performance. Additionally, while this study demonstrates the importance of long-term historical context and short-term agent-specific information, exploring adaptive mechanisms that dynamically adjust the emphasis between these dependencies could further improve the model's responsiveness to real-time changes.

\end{document}